# Time-Critical Dynamic Decision Making

Yanping Xiang    Kim-Leng Poh
Department of Industrial & Systems Engineering
National University of Singapore
Kent Ridge, Singapore 119260
{engp7600, isepohkl}@nus.edu.sg

## Abstract

Recent interests in dynamic decision modeling have led to the development of several representation and inference methods. These methods however, have limited application under time critical conditions where a trade-off between model quality and computational tractability is essential. This paper presents an approach to time-critical dynamic decision modeling. A knowledge representation and modeling method called *the time-critical dynamic influence diagram* is proposed. The formalism has two forms. The condensed form is used for modeling and model abstraction, while the deployed form which can be converted from the condensed form is used for inference purposes. The proposed approach has the ability to represent space-temporal abstraction within the model. A knowledge-based meta-reasoning approach is proposed for the purpose of selecting the best abstracted model that provide the optimal trade-off between model quality and model tractability. An outline of the knowledge-based model construction algorithm is also provided.

## 1 INTRODUCTION

The goal of dynamic decision making is to select an optimal course of action that satisfies some objectives in a time-dependent environment. The decisions may be made in different stages and each stage may varies in duration. A number of dynamic decision modeling formalisms have been proposed by various researchers. These include dynamic influence diagrams (DIDs) (Tatman and Shachter 1990), temporal influence diagrams (Provan 1993), Markov cycle trees (Beck and Pauker 1983), stochastic trees (Hazen 1992), and Dynamo (Leong 1994). Although these models provide relatively efficient methods for representing and reasoning in a time-dependent domain, the process of computing the optimal solution has remain intractable. A decision maker may have to spend a large amount of time in the modeling and solution processes that left very little or no time for the action to be carried out. This problem is particularly significant for large models involving temporal relations. Existing approaches to modeling and solving dynamic decision problems are therefore not appropriate for time-critical applications. Hence a more effective and practical approach to time-critical dynamic decision modeling is needed.

Time-critical dynamic decision problems have been discussed in several research communities. An important part of decision analysis is the formulation of the decision problem. Modeling time and the needs to deal with time-pressured situations are considered to be the greatest challenges in developing time-critical dynamic decision-support systems. We believe that a major reason for this perceived difficulty is the lack of modeling techniques that provide explicit support for the modeling of temporal processes, and for dealing with time-critical situations. In this paper, we propose a formalism called *time-critical dynamic influence diagrams* (TDID), that provide explicit support for the modeling and solution of time-critical dynamic decision problems.

In our approach we utilize the notion of abstraction to simplify the computational complexity of large and complex models. Previous research efforts on abstraction had mainly focused on data abstraction. For example, the KBTA (knowledge-based temporal abstraction) method (Shahar 1997) is a knowledge-based framework for the representation and application of the knowledge required for abstraction of high-level concepts from time-oriented data. Further research is needed on decision model abstraction methods and the selection of best situation-specific abstraction model. For a given domain, there exists a suite of possible decision models specified at different levels of space-temporal abstraction. Almost all previous research relied on the domain experts to assess the "goodness" of different abstractions. We adopt here the use of meta-reasoning to select an optimal model based on the best tradeoffs between decision quality and computational complexity.

This paper is organized as follows: In Section 2, we present the time-critical dynamic influence diagram used for the modeling and representation of time-critical dynamic decision problems. In Section 3, we show the applications of space and temporal abstractions using TDID. In Section 4, we propose a model-based approach to meta-reasoning for the selection of the best abstraction model and describe the model construction process. Finally, in Section 5, we conclude by summarizing and provide directions for further research.

## 2 TIME-CRITICAL DYNAMIC DECISION MODELING

### 2.1 TIME-CRITICAL DYNAMIC INFLUENCE DIAGRAM (TDID)

TDID is designed to facilitate the modeling and solution of time-critical dynamic decision problems. It extends standard influence diagram (Howard and Metheson 1981) by including the concepts of temporal arcs and time sequences. It also incorporates dynamic influence diagram (Tatman and Shachter 1990) as a representation for inference purposes.

A TDID model has two forms: the *condensed form* and the *deployed form*. Figure 1 shows an example of the condensed form. This form is mainly used to represent or define the dynamic decision model and is the form used in the modeling process. Figure 2 shows the deployed form for the same model. It is used only for inference purposes and is constructed from the condensed form. Although in principle both forms can be converted to and from each other, they serve different purposes in the modeling and solution processes.

Each node in a TDID represents a set of time-indexed variables. The set of time indices may be different from one node to another, but they must be subsets of a master time sequence. The arcs in a TDID are called temporal arcs and they denote both probabilistic and temporal (time-lag) relations among the variables. Solid arcs in the condensed form represent instantaneous probabilistic relations, while broken arcs represent time-lagged probabilistic relations. TDID allows for the coexistence of nodes of different temporal detail in the same model.

The TDID in Figure 1 has a master time sequence of <1,2,3,4>. Node $Y$ represents a set of chance variables indexed by this time sequence, whereas node $X$ represents a set of chance variables indexed by the subsequence <1, 3>. In this example, variable $X$ has been temporally abstracted by omitting its value at some intermediate time indices. This is evident from the deployed form in Figure 2 where nodes $Y_1, Y_2, Y_3$ and $Y_4$ are represented while only nodes $X_1$ and $X_3$ are probabilistically represented, and nodes $X_2$ and $X_4$ are assumed to be deterministically dependent (possibly equal) on $X_1$ and $X_3$ respectively. A formal definition of time-critical dynamic influence diagrams is given below.

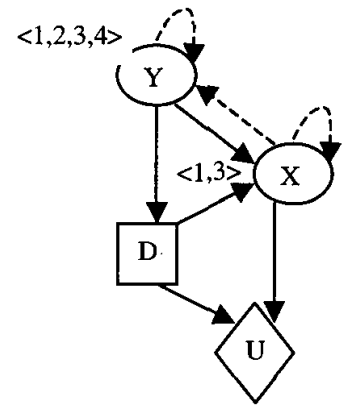

Figure 1: An Example Of A TDID In Condensed Form.

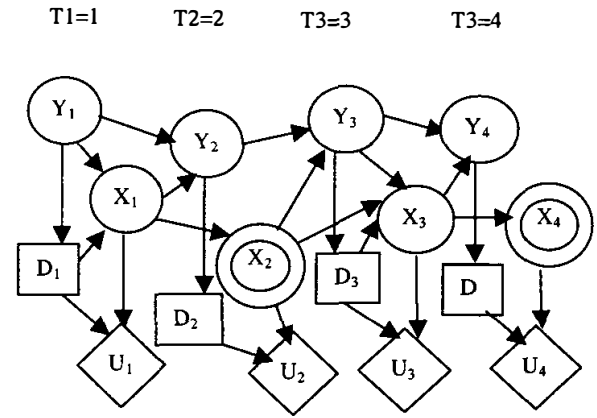

Figure 2: The Deployed Form Of The TDID In Figure 1

**Definition:** *The condensed form of time-critical dynamic influence diagrams* is a 7-tuple $<T_m, D, C, V, A_i, A_t, P>$ where

$T_m$ is a set of time indices called the *master time sequence*.

$D$ is a set *temporal decision variables*. Each $D \in D$ is a sequence of decision variables indexed by a time sequence $T_D \subseteq T_m$.

$C$ is a set of *temporal chance variables*. Each $C \in C$ is a sequence of chance variables indexed by a time sequence $T_C \subseteq T_m$.

V is a set of *utility functions* indexed by a time sequence $T_V \subseteq T_m$.

$A_i \subseteq (D \cup C) \times (D \cup C \cup \{V\})$ is a set of *instantaneous arcs* such that $\forall X \in (D \cup C)$ and $\forall Y \in (D \cup C \cup \{V\})$, $(X, Y) \in A_i$ if and only if there exists an instantaneous arc from node $X$ to node $Y$.

$A_t \subseteq (D \cup C) \times (D \cup C \cup \{V\})$ is a set of *time-lag arcs* such that $\forall X \in (D \cup C)$ and $\forall Y \in (D \cup C \cup \{V\})$, $(X, Y) \in A_t$ if and only if there exists a time-lag arc from node $X$ to node $Y$.

**P** is a set *conditional probability distributions*. For each chance node $X \in C$, we assess a sequence of conditional probability distributions $p(X_i \mid \pi(X_i))$ where $i \in T_X$, $\pi(X_i)$ is the set of nodes $Y_j$ such that $(Y, X) \in A_t$ and $j = \max \{ k \mid k \in T_Y, k < i \}$, or $(Y, X) \in A_i$ and $j = i$.

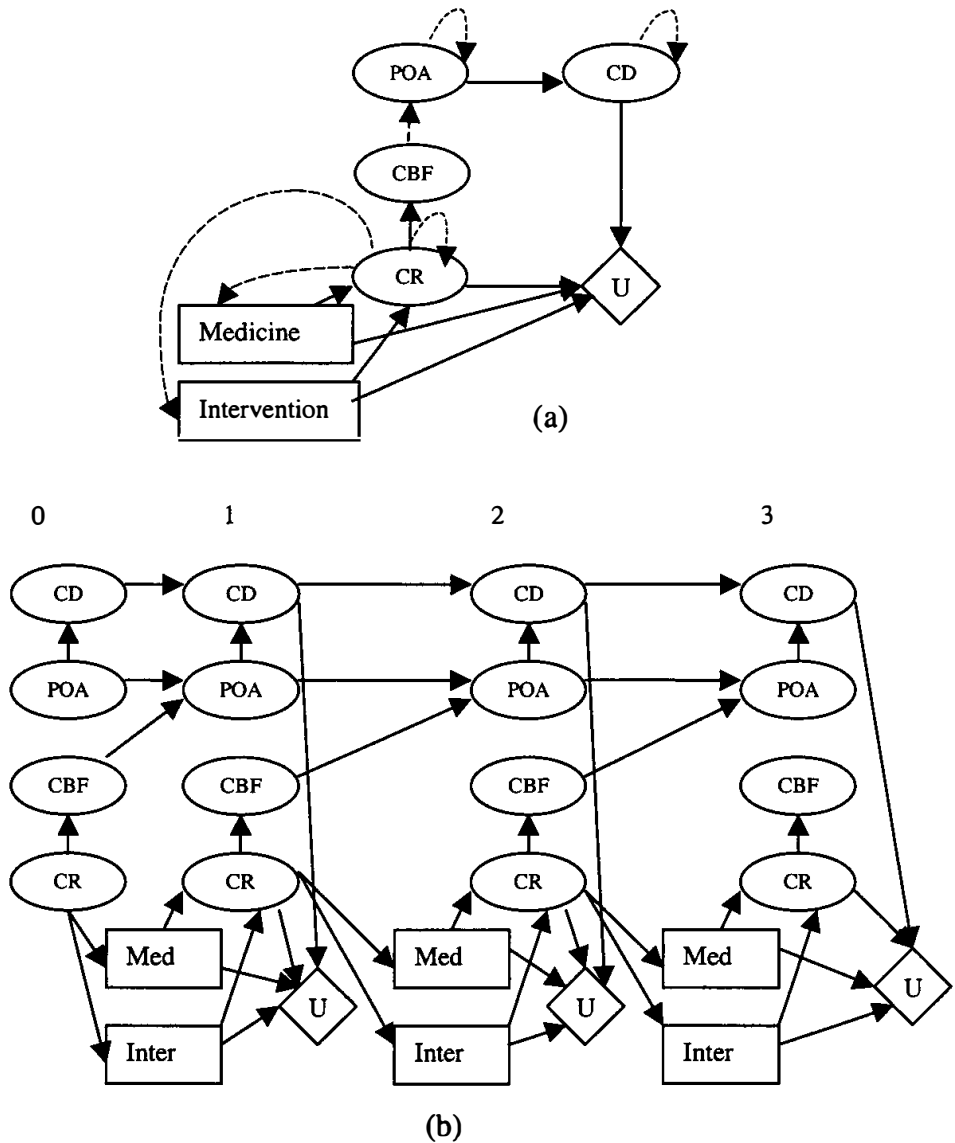

Figure 3: TDID For Cardiac Arrest Example

## 2.2 AN EXAMPLE

We shall use a medical example from the domain of cardiac arrest (Ngo *et al* 1997) to illustrate the use of TDID. In this problem, the goal of the medical treatment is to maintain life and to prevent anoxic injury to the brain. The observable variable is the electrocardiogram or rhythm strip (cr). While patient survival is of primary importance, cerebral damage must be taken into account and can be viewed as part of the cost in a resuscitation attempt. The length of time that patient has been without cerebral blood flow (cbf) determines the period of anoxia (poa). If the patient has ineffective circulation for more than five minutes, there is a likelihood of sustaining cerebral damage. This damage is persistent and its severity increases as the period of anoxia increases. Medical doctors treat a patient experiencing a cardiac arrest with a variety of interventions and medications.

The cardiac arrest problem may be represented by the TDID in condensed form as shown in Figure 3(a). We have assumed that the master time sequence is <1,2,3>, and all time sequences are the same as the master sequence. The real time between two time indices is 1 minute. Figure 3(b) shows the deployed form for the same model.

## 2.3 PROPERTIES OF TDID

We describe here the properties of TDID. First, we note that both forms of the TDID can be converted to and from each other. The condensed representation permits parsimonious descriptions of models and model abstraction. Inference using established algorithms for solving dynamic influence diagrams is then carried out by converting the condensed form to the deployed form and then adding a super value node.

TDID models time explicitly. It describes time in a clear and unambiguous manner. For example, in Figure 1, it is clear that $X$ is a possible cause of $Y$ and that this causal effect is delayed by 1 time unit. Through the use of temporal arcs and time sequences, a TDID has the ability to explicitly model how the underlying events evolve with time.

TDID allows for flexible temporal patterns. Compared with dynamic influence diagram in which temporal patterns of interest are predefined, TDID has temporal patterns that can be dynamically modified.

TDID provides a relatively high degree of reusability and modifiability of models. For instance, we may use the condensed form to describe a concise description of the domain, and it can then be easily reused. TDID also allows for dynamic modification of the model after the deployed form is produced from the condensed form.

Finally, TDID supports model abstraction. It provides a method to represent model abstraction, including different level of space abstraction and temporal abstraction. Details of model abstraction in TDID are given in Section 3.

## 2.4 CONVERTING THE CONDENSED FORM TO DEPLOYED FORM

The algorithm for converting a TDID in condensed form to its deployed form is as follows:

1. The time pattern for the deployed form is determined by master time sequence.
2. The graphical structure of the TDID without temporal-lag arc is replicated $N$ time, where $N$ is the number of time steps in the master time sequence. Let $ID_i$ be the $i$th influence diagram for $i=1,…, N$.
3. Connect the nodes in two different time slices according the temporal lag arc.
   **For** each the temporal arc **do**
       **For** $i = 1$ to $N$-1 **do**
           Add arc from the parent node in time slice $i$ to child node in time slice $i+1$;
4. Abstracting nodes
   **For** each node $X$ with time sequence $T_X \subset T_m$ **do**
       Partition $T_m$ into abstraction groups with members of $T_X$ as the starting index of each group.
       **For** each partition **do**
           The node indexed by the $T_X$ is the abstracted node. The other nodes in the same partition are assumed to be equal to the abstracted.
   Eliminate any barren nodes (Shachter 1986).
5. Insert probability distribution for expanded each node.

## 3 MODEL ABSTRACTION USING TDID

In this section we show how TDID provides a flexible, expressive and efficient formalism for representing model abstraction, including temporal abstraction and space abstraction. Abstraction of knowledge from domain experts provides high level building blocks that assists in both the development and maintenance of large knowledge-bases in decision-theoretic applications. Briefly, model abstraction is the task of creating context-sensitive interpretations of decision model in terms of higher-level space context and temporal patterns. The *input* to the model abstraction task is a set of abstraction goals and domain-specific abstraction knowledge. The *output* of the model abstraction task is a set of models at a higher level of space-temporal abstraction (Shahar 1997; Combi and Shahar 1997).

In our approach, the model abstraction task is decomposed into three sub-tasks: context interpretation,

space abstraction, and temporal abstraction. These tasks are supported by a domain knowledge base. Context interpretation is a set of relevant interpretation contexts, such as relation between a temporal pattern, the context of state in different abstraction levels and relations. Space abstraction focuses on abstracting the network within a time slice, including abstracting a group of state variables based on concept context, and summarising influence paths based on nodes reduction. Figure 4 shows an example of space abstraction of the TDID in Figure 3(a). This abstraction could be the result of a response to a massive myocardial infraction where we need not consider cerebral damage. The CD node and all subsequent barren nodes may be eliminated.

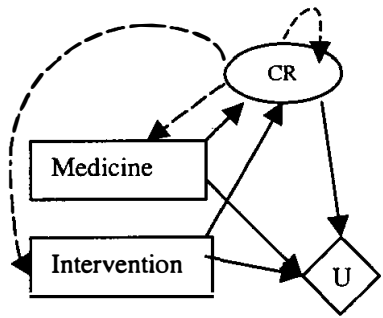

Figure 4: Space Abstraction Of The Model In Figure 3(a).

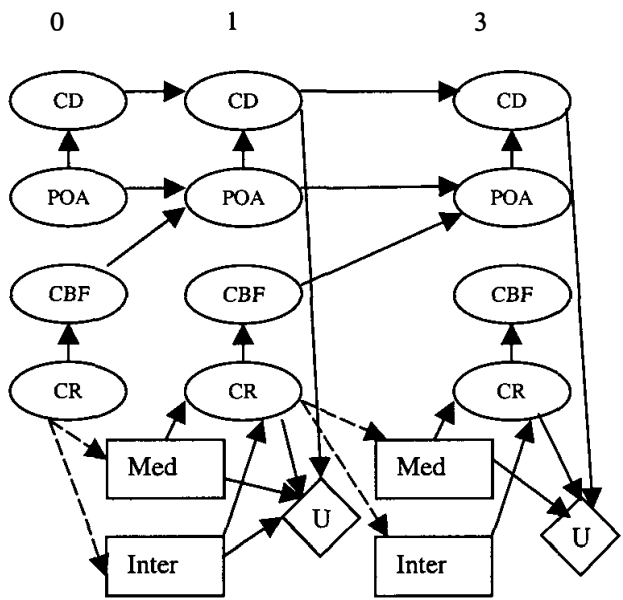

Figure 5: Temporal abstraction of the TDID in Figure 3(a).

Time abstraction may be performed on TDID by modifying the time sequences embedded in the model. Temporal abstraction in TDID leads to new context interpretations of the variables. In Figure 3, we assumed that the master time sequence is <1,2,3>, and all time sequences are the same as the master sequence. If all time sequences are abstracted to the subsequence <1, 3>, then the resulting time-abstracted TDID is shown in Figure 5 where the nodes in time slice 2 have been omitted.

## 4 MODEL SELECTION AND CONSTRUCTION

In the previous section, we showed how model abstraction may be applied in TDID. However, given a specific problem there exists many different possible space-temporal abstractions, and not all abstractions are equally good. Most of the previous research had recognized the usefulness of abstraction and but had relied on the domain experts to indicate the best level of abstraction. Here we address the problem of finding the best model among the set of possible space-temporal abstractions using meta-reasoning.

### 4.1 META REASONING

Meta-reasoning (Horvitz, 1990; Russell, 1991) enables a system to direct the course of its computations according to the current situation. In time critical applications, it is necessary that decision making effort be directed towards computation sequences that appear likely to yield good decisions. It is also important to consider tradeoffs between computational complexity and decision quality.

Figure 6 shows a decision-theoretic perspective of the meta-reasoning process. The goal is to determine the best abstraction model. For a specific problem, there exists a set of possible space-temporal abstraction models. We choose the best model based on trade-off between the computation cost and model quality. The computation cost is determined by computation time, while the model quality directly affects the quality of the action taken.

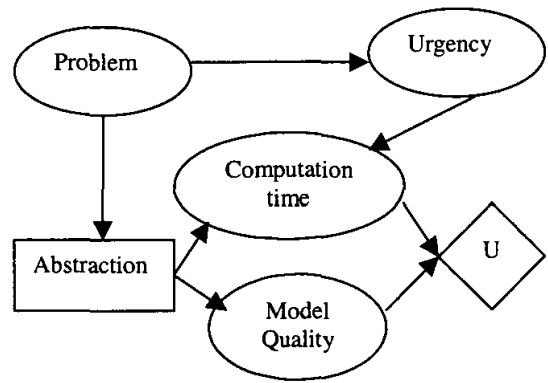

Figure 6: A Decision-theoretic perspective of meta-reasoning

We first consider model quality. For a given problem or situation, there exists a suite of models for solving the problem. Let $M$ be a set of different space-temporal abstraction models for the problem. For each model $m_i \in M$, let the maximum expected utility it yields be $u^*(m_i)$ which we will use as a measure of the model quality. Next, we consider the computation time. For a specific algorithm used to solve the model, the computation time $t$ can be estimated by assessing a function $C_t(S_i, N_i)$ where $S_i$ refers to the space complexity of the model $m_i$, and $N_i$ refers to the number of time intervals in model $m_i$.

We use the term *comprehensive value* $u_c(m_i)$ to refer to the overall utility that include both model quality and computational time. The comprehensive value is a function of the object-level utility $u_0$, and the inference-related cost, $u_i$ (Horvitz 1990). The *object-level utility* is the value associated with the information represented by the computed result of model $m_i$ without regard to the cost of reasoning. The *inference-related cost* is the penalty incurred while delaying action to arrive at the result. We define the comprehensive utility as the difference between the object-level utility $u_o(m_i) = u^*(m_i)$, and the inference-related cost, $u_i(m_i) = C_t(S_i, N_i)$. Hence we write:

$$u_c(u_o(m_i), u_i(m_i)) = u^*(m_i) - C_t(S_i, N_i)), \quad m_i \in M. \quad (1)$$

The best model is that model with maximum $u_c$.

### 4.2 A KNOWLEDGE-BASED APPROACH TO META-REASONING

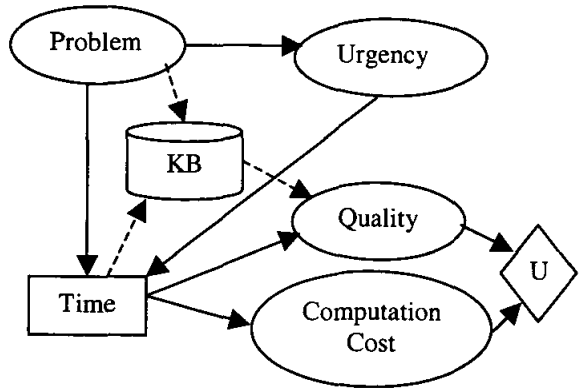

Figure 7: A Practical Knowledge-based Approach To Meta-reasoning.

The approach to meta-reasoning given in the previous subsection is intractable. We propose here a tractable approach to perform the meta-level analysis as shown in Figure 7. The goal of the meta-analysis is now to determine the length of time for computation, and then to identify the best model for doing so. We will use a knowledge base to support the process. We store a set of different space-temporal abstraction models in domain specified knowledge base and choose the "best" model based on the consideration of computation cost and quality.

Model quality may be approximated as follows: Suppose that the available computation time is $t$, and $M_{\leq t}$ is the set of models whose computational time is less than or equal to $t$, then model quality is

$$Q(M_{\leq t}) = \max u^*(m_i) \text{ subject to } m_i \in M_{\leq t}. \quad (2)$$

Let $m^* \in M_{\leq t}$ be the model whose maximum expected utility is equal to $Q(M_{\leq t}) = u^*(m^*)$, i.e., model $m^*$ has the highest quality among all models whose computation time is less than or equal to $t$.

By expending some quantity of reasoning resource $t$, e.g., computation time, model quality can be enhanced since the set of feasible models $M_{\leq t}$ is now larger. We define the comprehensive utility for a given computation time $t$ to be the difference between the object-level utility $Q(M_{\leq t})$ and the inference related cost $u_i(t)$. That is

$$u_c(Q(M_{\leq t}), t) = Q(M_{\leq t}) - u_i(t) \quad (3)$$

We define the net change in $u_c$ in return for an allocation of some computational resource to reasoning, as the *expected value of computation* (*EVC*) (Horvitz 1990). If $t_0$ is the amount of resources already committed, then the *EVC* for expending further resources $t$ is

$$EVC(t) = u_c(Q(M_{\leq t}), t) - u_c(Q(M_{\leq t_o}), t_o) \quad (4)$$

or

$$EVC(t) = [Q(M_{\leq t}) - Q(M_{\leq t_o})] - [u_i(t) - u_i(t_o)]$$

We can use the *EVC* to compare the value of extending the computation by different length of time and identify the ideal computation resource $t^*$ with the greatest *EVC*, i.e.,

$$t^* = \text{argmax } [EVC(t)]. \quad (5)$$

The best model is then the model $m^{**}$ such that

$$u^*(m^{**}) = Q(M_{\leq t^*}). \quad (6)$$

We observed that under time-critical conditions, meta-reasoning can provide useful control of the computational complexity by selecting the optimal abstraction model. In particular, it provides a straight forward way of incorporating flexibility to perform tradeoffs between the object-level value and the inference-related cost.

### 4.3 KNOWLEDGE-BASED MODEL CONSTRUCTION

We describe briefly here the model construction process, which is supported by a domain knowledge base. The steps for the model construction process are:

1. Given a time-critical dynamic decision problem, specify the problem requirements, such as its urgency and deadline.

2. Select a set of models that satisfies the requirements from the domain knowledge base.

3. Select the optimal model from the set of modes based on model quality and computation cost.

4. Modify or customize the TDID according to user requirements if necessary.

5. The TDID is converted into the deployed form, a super value node is added and the optimal solution determined.

## 5  SUMMARY AND CONCLUSION

In this paper, we have proposed an approach to time-critical dynamic decision making. The method is independent of any particular problem domain. A formalism for knowledge and model representation called the time-critical dynamic influence diagram was proposed. The TDID has two forms. The condensed form is used mainly for modeling and space-temporal abstraction, while the deployed is used only for inference purposes. We have shown how space and temporal abstraction may be carried out with TDID. In order to select the best abstracted model, we introduced the use of meta-reasoning, and proposed a practical knowledge-based approach to perform tradeoff between decision quality and computational complexity. Finally, we provide an outline of the knowledge-based model construction process.

Our work here is related to a number of previous work as well as some on going ones. The idea of representing a temporal sequence of probabilistic models into a compact form had been reported by Aliferis et. al. (1995, 1997). These related work had mainly focused on bayesian networks while our work here include temporal decisions. The use of knowledge base to support temporal abstraction of data had been investigated by Shahar and Musen (1996). The use of meta-reasoning for directing the course of computation have been investigated by Horvitz (1990) and Russell (1991). The idea of using expected value of refinement and value of computation to direct model refinement and abstraction was briefly described in Poh and Horvitz (1993).

Finally, the authors are presently working on the application of the approach to a time-critical medical domain involving head injury critical care in a hospital in Singapore.

### Acknowledgments

The authors would like to thank the Tze-Yun Leong, David Harmanec and other research group members for their helpful suggestions and comments on this work. This work is partly supported by a strategic research grant from the Singapore National Science and Technology Board and Ministry of Education. Yanping is supported by a National University of Singapore research scholarship.